\documentclass{article}
\usepackage{ijcai26}

\usepackage{times}
\usepackage{soul}
\usepackage{url}
\usepackage[hidelinks]{hyperref}
\usepackage[utf8]{inputenc}
\usepackage[small]{caption}
\usepackage{graphicx}
\usepackage{amsmath}
\usepackage{amsthm}
\usepackage{booktabs}
\usepackage{algorithm}
\usepackage{algorithmic}
\usepackage[switch]{lineno}
\usepackage{amssymb}

\newtheorem{theorem}{Theorem}

\title{SPOTting the Future: Lookahead Explanations for Deep Reinforcement Learning}
\author{
Tamar Gozlan$^1$\and
Claudia Goldman$^2$\\
\affiliations
$^1$Benin School of Computer Science and Engineering,
The Hebrew University of Jerusalem\\
$^2$Hebrew University Business School, Data Science Department,
The Hebrew University of Jerusalem\\
\emails
\{tamar.gozlan, claudia.goldman\}@mail.huji.ac.il}
\date{}

\begin{document}
\maketitle

\begin{abstract}
Deep reinforcement learning (DRL) agents achieve strong performance in complex environments, yet their decision-making processes remain difficult to interpret. We introduce \textbf{SPOT} (Sampling Policy Observation Tree), a novel model-agnostic, sampling-based framework for interpreting DRL policies. Given access to the policy and an environment simulator, SPOT constructs an interpretable finite-horizon tree by sampling actions and recursively simulating the resulting successor states. The tree provides an empirical representation of the policy's action preferences and their possible downstream evolution.
We provide formal guarantees establishing SPOT's asymptotic recovery of the policy's unique most probable action and characterizing its disagreement behavior under high-entropy policies.
We demonstrate SPOT in the SUMO-RL traffic-signal control domain. The case study illustrates how its tree-based representation can be used to inspect policy preferences, compare alternative future trajectories, and reveal downstream behaviors that are not visible through single-timestep feature-attribution methods.
\end{abstract}

\section{Introduction}
Despite its success in complex decision-making tasks, the application of deep reinforcement learning (DRL) in real-world settings remains limited due to its reliance on large amounts of data, sensitivity to changing environments, and lack of interpretability (e.g., \cite{Lekadir2025}).
Existing explainable AI (XAI) methods for DRL largely focus on feature-level attribution, identifying which inputs influence a given action. While useful, these approaches are inherently limited to single timesteps and do not capture how decisions affect future states. As a result, they fail to explain long-term consequences or detect suboptimal decisions whose impact emerges over time.

We introduce \textbf{SPOT} (Sampling Policy Observation Tree), a novel XAI framework that addresses these limitations by modeling the distribution of future trajectories induced by the agent’s policy. SPOT constructs an interpretable tree through action sampling and state expansion, enabling trajectory-aware explanations that explicitly compare the outcomes of alternative actions.

We consider a human-in-the-loop setting where the agent acts as a decision-support system. Explanations must therefore be actionable, enabling operators to assess whether to follow or override recommendations. SPOT provides domain-readable summaries that highlight key signals to support such decisions.

We demonstrate SPOT in a SUMO-RL traffic control environment, showing that it uncovers behaviors missed by single-timestep explanations and enables more informed decisions.

\section{Related Work}

Deep reinforcement learning (DRL) combines reinforcement learning \cite{sutton2018rl} with deep neural networks to learn policies from high-dimensional observations. This integration allows RL to scale to complex environments by leveraging powerful function approximation capabilities \cite{dong2020drl,arulkumaran2017drl}.

\subsection{Taxonomy of DRL Methods}

DRL methods are commonly categorized into value-based, policy-based and actor-critic methods. This classification reflects different approaches to solving Markov Decision Processes (MDPs), where the goal is to maximize expected cumulative reward \cite{dong2020drl}. The MDP is given by the tuple $<S,A,P,R,\gamma>$, where $S$ represents the state space, $A$, the actions set, $P$ the transition probabilities (that are not explicitly known by the DRL solution), the reward function $R$ and a discount parameter $\gamma$.

\paragraph{Scope of Taxonomy.}
Our taxonomy focuses on \emph{pure DRL} methods, where the core learning and decision-making components are parameterized by deep neural networks. Hybrid systems are excluded because their additional mechanisms fall outside a unified DRL taxonomy. However, our explainability framework can still be applied to the DRL component within such systems.

\begin{figure*}[t]
    \centering
    \includegraphics[width=0.7\textwidth]{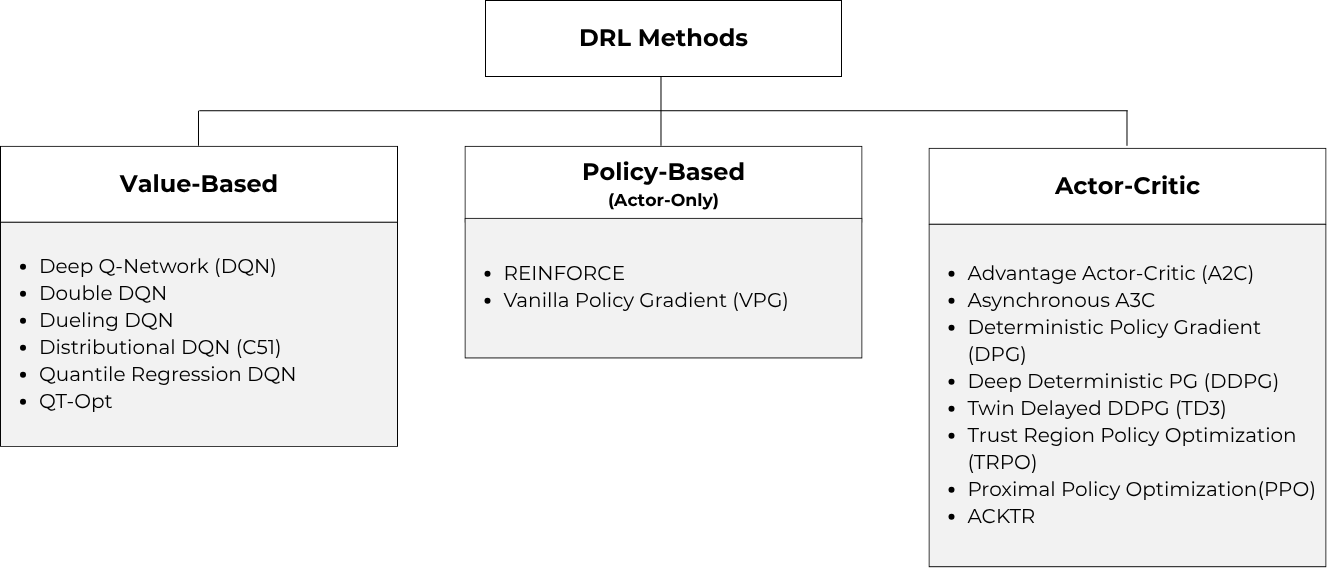}
    \caption{Taxonomy of deep reinforcement learning methods, categorized into value-based, policy-based, and actor-critic approaches. Based on \protect\cite{dong2020drl}.}
    \label{fig:drl-taxonomy}
\end{figure*}

\subsection{Value-Based Methods}

Value-based reinforcement learning methods aim to estimate value functions that quantify the expected return of states or state-action pairs. The state-value function $V(s)$ represents the expected cumulative reward when starting from state $s$, while the action-value function $Q(s,a)$ represents the expected return of taking action $a$ in state $s$ and following a given policy thereafter.

\noindent
Formally, the state-value function, following a policy $\pi$, is defined as:
\[
V^\pi(s) = \mathbb{E}_\pi \left[ \sum_{t=0}^{\infty} \gamma^t r_t \mid s_0 = s \right],
\]
and the action-value function as:
\[
Q^\pi(s,a) = \mathbb{E}_\pi \left[ \sum_{t=0}^{\infty} \gamma^t r_t \mid s_0 = s, a_0 = a \right].
\]

Q-learning forms the foundation of modern value-based reinforcement learning, where the optimal policy is obtained by selecting actions that maximize the learned $Q$-function. Most subsequent advancements build upon this core formulation, introducing improvements in stability and performance without fundamentally altering the underlying algorithm \cite{mckenzie2022value}.

A major breakthrough occurred with the introduction of the Deep Q-Network (DQN), which combines Q-learning with deep neural networks to approximate value functions in high-dimensional state spaces \cite{mnih2013dqn}. DQN enabled end-to-end learning from raw inputs and achieved human-level performance on Atari benchmarks. Following this milestone, a rich line of work has emerged extending DQN, with each variant designed to address specific shortcomings such as overestimation bias, inefficient sampling, or representation limitations. Notable examples include Double DQN \cite{vanhasselt2016double}, prioritized experience replay, and dueling architectures. While these methods differ in their design objectives, they largely retain the core Q-learning framework.

\subsection{Policy-Based Methods}

Policy-based methods directly parameterize the policy $\pi(a|s)$ and optimize it using gradient-based methods. The theoretical foundation of policy gradient methods was established by the REINFORCE algorithm \cite{williams1992reinforce}, which provides an unbiased estimator of the gradient of expected reward.

\[
\nabla_\theta J(\theta) = \mathbb{E}\left[(r - b)\nabla_\theta \log \pi_\theta(a|s)\right]
\]

This expression defines how the policy parameters $\theta$ should be updated to maximize expected return. The term $\nabla_\theta \log \pi_\theta(a|s)$ captures the sensitivity of the policy to parameter changes, indicating how a small adjustment in $\theta$ affects the probability of selecting action $a$ in state $s$. The scalar term $(r - b)$ represents the advantage of the action, measuring how much better or worse the observed return $r$ is compared to a baseline $b$.

Intuitively, this update rule increases the probability of actions that yield higher-than-expected returns and decreases the probability of actions that perform worse than expected. Using the baseline $b$ reduces the variance of the gradient estimate, leading to more stable learning.

In practice, modern implementations of policy gradient methods are often referred to as Vanilla Policy Gradient \cite{schulman2017ppo}, which follows the same fundamental principle as REINFORCE but typically incorporates practical improvements such as batch updates, reward-to-go estimation, and normalization techniques to improve training stability and efficiency.

A key insight of policy gradient methods is that they perform stochastic gradient ascent on the expected return without requiring an explicit model of the environment. In expectation, the update direction aligns with the true gradient of the objective function, ensuring that the policy improves over time.

\subsection{Actor-Critic Methods}

Actor-critic methods combine the advantages of value-based and policy-based approaches by maintaining two components: an \textit{actor} that represents the policy, and a \textit{critic} that estimates a value function \cite{grondman2012actorcritic}. These approaches jointly learn a policy and a value function, unlike earlier methods that learn only one.
The critic evaluates the current policy by estimating the expected return, while the actor updates the policy parameters based on this feedback. This interaction reduces the high variance inherent in policy gradient methods while retaining their favorable convergence properties.

\noindent
Typically, the policy update is guided by the temporal-difference (TD) error which serves as an estimate of the advantage of the taken action:
\[
\delta_t = r_{t+1} + \gamma V(s_{t+1}) - V(s_t),
\]

The resulting policy gradient update takes the form:
\[
\nabla_\theta J(\theta) = \mathbb{E}\left[\delta_t \nabla_\theta \log \pi_\theta(a|s)\right],
\]
where $\nabla_\theta \log \pi_\theta(a|s)$ captures the sensitivity of the policy to parameter changes, and $\delta_t$ indicates whether the action performed better or worse than expected.

As in policy-based methods, this update reinforces actions yielding higher-than-expected returns while suppressing suboptimal ones. Using a learned value function as a baseline reduces the variance of the gradient estimates, leading to more stable and efficient learning.

\subsection{Explainable Deep Reinforcement Learning Methods}

Explainability in deep reinforcement learning (XRL) aims to make an agent's decisions, strategies, and failure modes more transparent to human observers. This need is particularly critical in DRL, where policies are typically represented by deep neural networks, making the decision-making process difficult to interpret in complex environments. Recent work categorizes XRL methods according to the level at which explanations are generated, including feature-level, state-level, dataset-level, and model-level approaches \cite{cheng2025surveyxrl}.

\paragraph{Feature-level explanations.}
Feature-level methods aim to identify which components of the input most influence the agent’s decisions. For example, SHAP \cite{lundberg2017shap} provides a model-agnostic framework based on Shapley values to attribute importance to input features. While these methods offer insight into which features influence decisions, they remain limited to single timesteps and primarily capture low-level correlations, without explaining how decisions affect future outcomes.

\paragraph{State-level explanations.}
State-level methods explain decisions by identifying important states in the agent’s trajectory and analyzing their impact on performance. 
Prior work includes offline methods that analyze pre-collected trajectories, such as AIRS \cite{yu2023airs}. These methods depend on the quality of collected data and may fail to capture unseen states. Online approaches, such as LazyMDP \cite{jacq2022lazymdp}, identify critical steps by modifying actions during interaction. Another line of work uses causal models to explain decisions; for example, \cite{madumal2019causal} learn structural causal models during training to capture cause-effect relationships between variables and actions, while assuming a directed acyclic graph (DAG) specifying the causal relationships between variables is given. These models enable the generation of contrastive explanations (e.g., “why this action and not another”) by reasoning over counterfactual scenarios. 
While effective for highlighting influential states, these approaches may not generalize well to long-term or global policy behavior.

\paragraph{Dataset-level explanations.}
Dataset-level methods explain agent behavior by quantifying how training data influences the learned policy. These approaches provide a global perspective by identifying influential trajectories or experiences. For example, Data Shapley methods \cite{ghorbani2019shapley} assign importance scores to data points based on their contribution to policy performance. While such methods help diagnose training dynamics and guide data collection, they are computationally expensive and do not provide insight into individual decisions.

\paragraph{Model-level explanations.}
Model-level methods provide structured, high-level explanations by analyzing the learned policy or the underlying environment model. For example, \cite{finkelstein2022model} generate explanations by applying transformations to a Markov Decision Process (MDP) until the agent’s behavior aligns with an observer’s expectations. These transformations reveal which aspects of the environment, such as constraints or dynamics, explain discrepancies between expected and observed behavior. While such approaches provide global insight into policy behavior, they often require access to a model of the environment and rely on additional assumptions (e.g., a human-observer policy).

\paragraph{Summary.}
Overall, existing XRL methods differ in both \emph{what} they explain and \emph{how} they explain it. Feature- and state-level methods are often useful for local, post-hoc interpretation, whereas dataset-level methods summarize broader behavior, and model-level methods provide more structured and often more semantically meaningful accounts of decision making.

\section{SPOT: Sampling Policy Observation Tree}

We propose \emph{SPOT} (Sampling Policy Observation Tree), an online sampling-based structure designed to expose and interpret the decision-making process of deep reinforcement learning agents. SPOT constructs a tree rooted at the current state and expands it using samples drawn from the agent’s policy, thereby providing an interpretable representation of action preferences and their relative likelihoods.

\subsection{Model-Agnostic Policy Extraction}

SPOT assumes access to a stochastic policy $\pi(a \mid s)$ defined over a finite action space $\mathcal{A} = \{a_1, \dots, a_k\}$. For policy-based and actor-critic methods, this policy is directly available as the output of the actor network. For value-based methods, which estimate an action-value function $Q(s,a)$, we derive a stochastic policy using a softmax transformation~\cite{sutton2018rl}:
\[
\pi(a_i \mid s) = \frac{\exp\left(Q(s,a_i)\right)}{\sum_{j=1}^k \exp\left(Q(s,a_j)\right)}
\]

\noindent
Since a stochastic policy can be derived for each class of deep reinforcement learning methods, SPOT is model-agnostic and can be applied in all settings.

\subsection{Tree Construction}

SPOT assumes access to an agent's policy \(\pi\) and an environment simulator
\(\mathcal{E}\) capable of generating successor states. Given a current state
\(s\), SPOT initializes a root node representing \(s\) and draws \(N\)
independent action samples:
\[
A_1,\ldots,A_N \sim \pi(\cdot\mid s).
\]

Let \(\mathcal{A}=\{a_1,\ldots,a_k\}\) denote the action space. Rather than
creating a separate branch for every sample, SPOT aggregates identical actions.
The visit count of action \(a_i\) is
\[
C_i
:=
\sum_{n=1}^{N}\mathbf{1}\{A_n=a_i\},
\qquad
\sum_{i=1}^{k}C_i=N.
\]

For each action \(a_i\) with \(C_i>0\), the simulator generates the successor
state
\[
s'_i=\mathcal{E}(s,a_i).
\]
SPOT then creates a child node representing \(s'_i\) and labels the
corresponding branch with \(a_i\) and \(C_i\). 
The procedure is applied recursively. At each node of depth \(d<K\), SPOT
samples \(N\) actions from the policy at the node's state, aggregates the
samples by action, and simulates the corresponding transitions. Expansion
continues until the predefined maximum depth \(K\) is reached, producing a
finite-horizon tree of the policy's local decision behavior. Algorithm~
\ref{alg:spot} summarizes the construction procedure.

\begin{algorithm}[t]
\caption{SPOT Tree Construction}
\label{alg:spot}
\begin{algorithmic}[1]

\REQUIRE Initial state $s$, policy $\pi$, simulator $\mathcal{E}$
\REQUIRE Sample count $N$, maximum depth $K$
\ENSURE SPOT tree $\mathcal{T}$ rooted at $s$

\STATE Initialize $\mathcal{T}$ with a root node representing $s$

\FOR{$d=0,\ldots,K-1$}
    \FOR{each node $v$ at depth $d$}
        \STATE Let $s_v$ denote the state represented by $v$
        \STATE Draw $A_1,\ldots,A_N \sim \pi(\cdot\mid s_v)$

        \FOR{each distinct sampled action $a_i$}
            \STATE $C_i \gets
            \sum_{n=1}^{N}\mathbf{1}\{A_n=a_i\}$
            \STATE $s'_i \gets \mathcal{E}(s_v,a_i)$
            \STATE Add a child representing $s'_i$ to $v$
            \STATE Label the branch with $(a_i,C_i)$
        \ENDFOR
    \ENDFOR
\ENDFOR

\STATE \textbf{return} $\mathcal{T}$

\end{algorithmic}
\end{algorithm}

\noindent
This expansion is applied recursively. At each node of depth $d$, we draw $N$ samples from the policy at the corresponding state and construct its child nodes accordingly. The process continues until a predefined maximum depth $K$ is reached, yielding a multi-level tree that captures the agent’s decision-making over a finite lookahead horizon.

\subsection{Theoretical Guarantees}

We provide theoretical justification for SPOT by analyzing the relationship between its empirical estimates and the underlying policy.

\begin{theorem}[Consistency with the Greedy Policy]
Let \(a^\star = \arg\max_{a \in \mathcal{A}} \pi(a \mid s)\) be the unique greedy action, and let \(C_i\) denote the visit count of action \(a_i\) obtained from \(N\) i.i.d. samples drawn from \(\pi(\cdot \mid s)\). Then,
\[
\Pr\!\left( \arg\max_{a_i \in \mathcal{A}} C_i = a^\star \right)
\to 1 \quad \text{as } N \to \infty.
\]
\end{theorem}

\noindent
This result establishes that SPOT recovers the greedy action in the limit of large sample sizes (proof in the Supplementary Material).

\vspace{0.5em}

\begin{theorem}[Disagreement Under Maximum Uncertainty]
Let \(\mathcal{A} = \{a_1,\dots,a_k\}\), and define
\[
T_N := \arg\max_{a_i \in \mathcal{A}} C_i,
\]
with ties broken uniformly at random. As the policy approaches the uniform distribution,
\[
\pi(a_i \mid s) \to \frac{1}{k},
\]
the disagreement probability satisfies
\[
\Pr(T_N \neq a^\star) \to \frac{k-1}{k}.
\]
\end{theorem}

\noindent
This result characterizes the behavior of SPOT under high-entropy policies, where action preferences become indistinguishable (proof in the Supplementary Material).

\subsection{Extending SPOT for Rich Policy Analysis}

While SPOT represents action preferences through visit counts, the tree structure naturally supports the integration of additional signals that enrich the interpretability of the agent’s decision-making process.

\paragraph{Node-Level Information.}
Each node in the SPOT tree can store auxiliary quantities beyond visit counts, including the critic’s value estimate $V(s)$, action-value estimates $Q(s,a)$, advantage estimates $A(s,a)$, and observed rewards along sampled trajectories.

These quantities are computed during tree expansion through environment interaction. For each sampled action, the environment is advanced from the node’s state to obtain a successor state $s'$ and an immediate reward $r(s,a)$. When a value function is available in the DRL model, the value estimates $V(s)$ and $V(s')$ are obtained directly from this function. The advantage $A(s,a)$ can be computed using a one-step temporal-difference estimate.

Incorporating these signals provides a richer characterization of each decision, capturing not only action likelihoods but also their expected utility and contribution to long-term outcomes.

\paragraph{Tree-Based Analysis.}
The SPOT structure enables the extraction of explanations from the agent’s decision-making process by analyzing the information encoded in the tree. Similar to planning-based approaches, where tree structures expose the agent’s reasoning over possible future trajectories, SPOT provides access to both local decisions and forward-looking behavior.

In particular, the explanation methods introduced by \cite{goldman2025realtime} are directly applicable in the SPOT setting. \emph{Plan Next} explanations describe the agent’s expected future behavior by identifying high-probability trajectories in the tree, effectively indicating what the agent is likely to do next. \emph{Contrastive} explanations compare alternative branches to explain why a selected action is preferred over other feasible actions, highlighting the differences between competing choices. Finally, \emph{Value Component} explanations decompose value-related quantities stored in the tree, such as accumulated rewards or value estimates, to explain the contribution of different factors to the agent’s decision, when such quantities admit a meaningful decomposition.

By supporting these explanation methods, SPOT enables a structured analysis of both the immediate and long-term consequences of decisions for DeepRL solutions. 

\paragraph{Global Behavior Summarization.}
Beyond local explanations, the SPOT framework can be used to construct a global view of an agent’s behavior by identifying and analyzing \emph{critical states}. Prior work has shown that states in which different actions lead to substantially different outcomes are particularly informative for understanding an agent’s strategy \cite{amir2018highlights}.

In SPOT, critical states can be identified by measuring the divergence between the outcomes of alternative actions. States where different actions lead to large differences in expected return or future trajectories are flagged as high-impact decision points. For example, a state where one action leads to low congestion while another results in significant delay is considered critical. To avoid redundancy, we select critical states that represent different regions of the state space (e.g., different traffic patterns or congestion levels). Aggregating the SPOT trees rooted at these states provides a structured summary of the agent’s policy; this component is not implemented in the current work and will be reported separately.

\section{Case Study: Traffic Signal Control}

In this section, we demonstrate SPOT on traffic-signal control environment built with SUMO~\cite{behrisch2011sumo}.  The underlying decision-making problem is to dynamically select the traffic signal phase at each timestep so as to minimize cumulative vehicle waiting time at a signalized intersection. In our setting the agent does not act autonomously. Instead, a human operator remains in the loop and uses the agent’s recommendations to guide decision-making.
In this context, explanations are not merely descriptive but must support actionable oversight: the operator must be able to assess whether to follow or override the agent’s decision. We show that SPOT produces temporally grounded, domain-readable explanations that surface critical operational signals that are invisible to single-timestep attribution methods like SHAP \cite{lundberg2017shap}. 
We first describe the explanation extraction pipeline (Section~\ref{sec:sumo_pipeline}), then the experimental setup (Section~\ref{sec:sumo_setup}), and finally a disruption scenario designed to expose the blind spot of attribution-only methods (Section~\ref{sec:sumo_scenario}).

\subsection{Explanation Extraction Pipeline}
\label{sec:sumo_pipeline}

Given a SPOT tree for each decision step, the explanation pipeline converts raw numerical tree data into human-readable text through four successive stages.

\paragraph{Stage 1 --- Observation decoding.}
The 21-dimensional observation vector produced by SUMO-RL is decoded into domain concepts over eight incoming lanes, including the active signal phase, a minimum-green indicator, and per-lane traffic features. (A detailed description of these components is provided in the Supplementary Material).

\paragraph{Stage 2 --- Path extraction and trend detection.}
Two paths are extracted from each tree. The \emph{chosen path} follows the highest-visit child at every depth, tracing the most likely future trajectory induced by the policy. This corresponds to the \emph{Plan-Next (PN)} explanation in Goldman et al. [\citeyear{goldman2025realtime}], capturing what the agent is expected to do next. The \emph{counterfactual path} starts from the depth-1 node with the second-highest visit count and then follows the highest-visit branch thereafter, representing the most plausible alternative trajectory. This corresponds to the \emph{Contrastive (CON)} explanation in Goldman et al. [\citeyear{goldman2025realtime}], capturing what would have happened had a different action been taken.

\noindent
For each path, we extract six scalar sequences along the trajectory:
critic values, immediate rewards, advantages, policy confidence,
per-lane queue lengths, and per-lane densities. For each sequence, we compute a qualitative trend label through the following steps:

\begin{enumerate}
    \item \textbf{Net change.} 
    We summarize each sequence by its overall change between the first and last timestep:
    \[
    \Delta = x_T - x_0.
    \]
    This captures whether the quantity increases, decreases, or remains stable along the path. 
    To interpret the magnitude of this change, we compare $|\Delta|$ to two thresholds scaled by the root critic value $V_0$:
    \[
    \delta_{\text{small}} = 0.2|V_0|,
    \qquad
    \delta_{\text{large}} = 0.6|V_0|.
    \]
    Changes smaller than $\delta_{\text{small}}$ are treated as negligible (\emph{flat}), 
    changes between $\delta_{\text{small}}$ and $\delta_{\text{large}}$ are considered \emph{small}, 
    and changes larger than $\delta_{\text{large}}$ are considered \emph{large}. 
    Combining magnitude with the sign of $\Delta$ yields five coarse categories:
    \emph{large positive}, \emph{small positive}, \emph{flat}, \emph{small negative}, and \emph{large negative}.

    \item \textbf{Trajectory shape.}
    We then examine step-by-step changes along the sequence to determine whether the trajectory is:
    (i) monotonic (consistently increasing or decreasing), or
    (ii) non-monotonic (changing direction).

    \item \textbf{Final classification.}
    Combining the net change and the trajectory shape produces one of seven trend categories:
    \emph{monotone-large-positive}, \emph{monotone-small-positive}, \emph{monotone-large-negative}, \emph{monotone-small-negative}, \emph{non-monotone-net-positive}, \emph{non-monotone-net-negative}, or \emph{flat}.
\end{enumerate}

\paragraph{Stage 3 --- Signal-strength filtering.}
Not all detected trends are informative enough to be included in the explanation. 
We therefore apply a signal-strength filter that retains only trends that exhibit 
sufficient magnitude or clear directional structure, and suppresses weak or ambiguous signals.

The filter removes noise (small fluctuations or inconsistent patterns) 
and keeps only trends that provide actionable insight into the system's behavior. 
Table~2 in the Supplementary Material summarizes the filtering criteria.

\paragraph{Stage 4 --- Stance scoring and actionable summary.}
The chosen path (\emph{chosen\_dir}) is compared against the alternative path (\emph{alt\_dir}) by aggregating evidence across multiple aspects into a single scalar score. Each aspect contributes a signed vote, where positive values support \emph{chosen\_dir} and negative values support \emph{alt\_dir}. The final score is obtained by summing all votes and is mapped to a recommendation: high positive scores indicate \textsc{Trust}, intermediate scores indicate \textsc{Watch}, and strongly negative scores indicate \textsc{Intervene}. Full scoring rules and thresholds are provided in the Supplementary Material file.

\paragraph{Actionable summary.}
At each decision step, the system produces a concise, one-paragraph summary for the operator. 
It includes a stance label (\textsc{Trust}, \textsc{Watch}, or \textsc{Intervene}), the key signals driving the decision, 
and an indication of the agent’s confidence. When the stance is \textsc{Intervene}, the summary also identifies the preferred alternative phase and explains why it is favored.

\subsection{Experimental Setup}
\label{sec:sumo_setup}

\begin{figure}[h]
    \centering
    \includegraphics[width=0.7\linewidth]{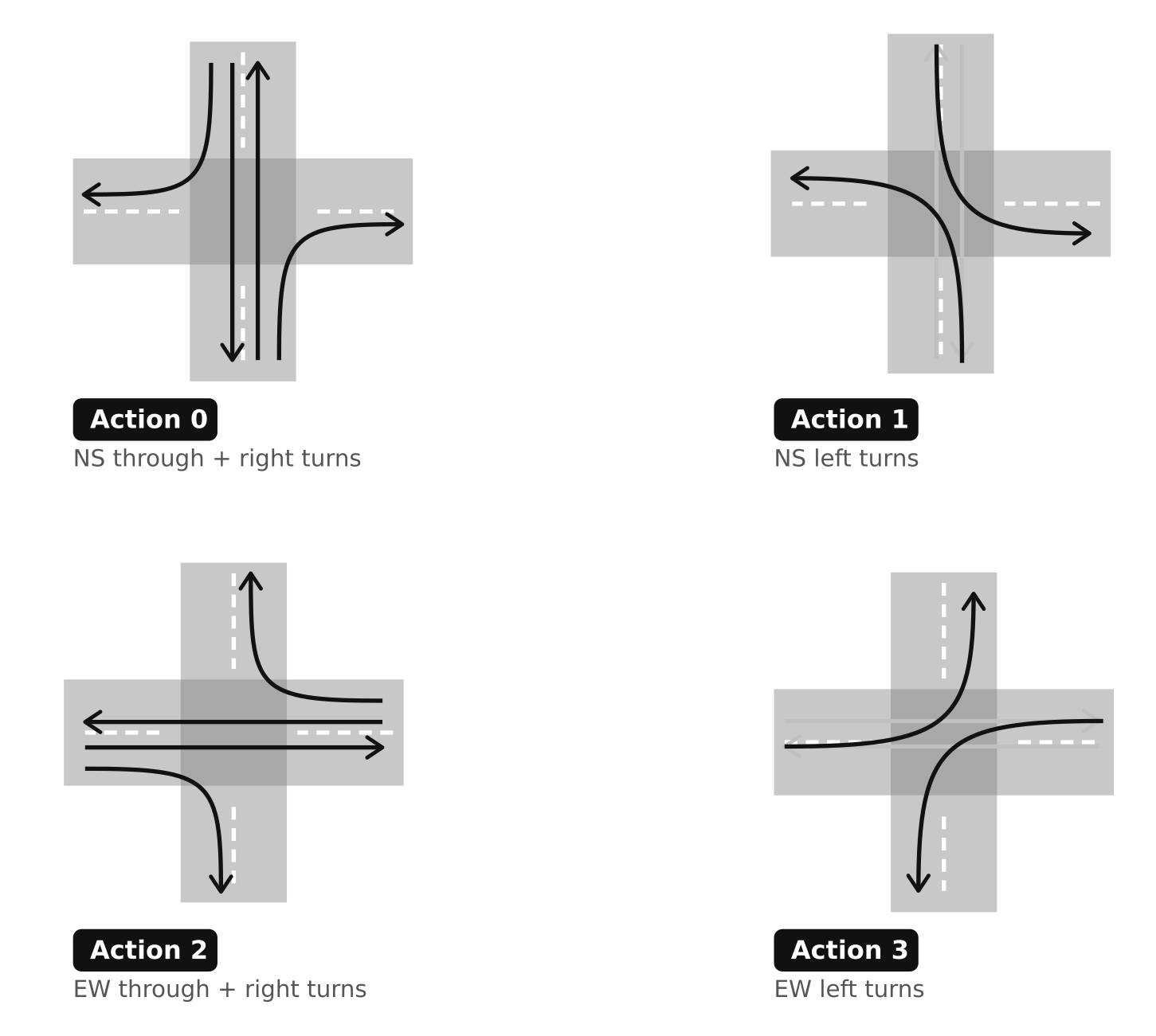}
    \caption{
    Signal phases (agent's actions) used in the SUMO-RL environment. 
    }
    \label{fig:sumo_phases}
\end{figure}

\paragraph{Simulation environment.}
We use the SUMO-RL Gymnasium wrapper\footnote{\url{https://github.com/LucasAlegre/sumo-rl}} with a single four-way signalized intersection. Each episode runs for 3{,}600 simulated seconds. The observation space follows the default SUMO-RL encoding:
\[
\begin{aligned}
\texttt{obs} = [&\texttt{phase\_one\_hot},\; \texttt{min\_green},\\
               &\texttt{lane densities},\; \texttt{lane queues}]
\end{aligned}
\]
which in our network yields a 21-dimensional state vector. The reward is the change in cumulative vehicle waiting time (\texttt{diff-waiting-time}), so negative reward corresponds to increasing delay. The action space consists of four discrete signal phases (Figure~\ref{fig:sumo_phases}),
each enabling a set of non-conflicting traffic movements. The minimum green time is 5\,s, and every phase transition incurs a 2\,s yellow interval. Consequently, the agent acts every 5 simulated seconds (\texttt{delta\_time}=5).

\paragraph{Policy.}
We train a Proximal Policy Optimization (PPO) agent for \(100{,}000\)
timesteps using Stable-Baselines3\footnote{\url{https://stable-baselines3.readthedocs.io/}}
with standard hyperparameters, including a discount factor of
\(\gamma=0.99\) and a clipping range of \(0.2\). The agent uses an MLP
actor--critic policy and is trained in the single-agent SUMO-RL
environment defined by the two-way single-intersection network and its
associated traffic-demand route file. At inference time, the agent acts
deterministically according to
\[
a_t=\arg\max_a \pi(a\mid s_t).
\]

\paragraph{SPOT Configuration.}
At each decision step, SPOT constructs a lookahead tree by sampling
\(N=100\) stochastic actions from the trained policy at every node and
expanding the tree to depth \(K=3\).

\subsection{Accident Scenario: Demonstrating SPOT’s Future-Aware Reasoning}
\label{sec:sumo_scenario}
\paragraph{Motivation.}
Methods such as SHAP provide explanations based solely on the current observation. As a result, disruptions that do not immediately affect the observed state—but lead to future degradation—remain undetected. To highlight this limitation, we design a scenario in which a lane blockage is placed upstream of the sensor range. This ensures that the current feature vector appears normal, while the system’s performance deteriorates over time.

\begin{figure}[h]
\centering
\includegraphics[width=0.6\linewidth]{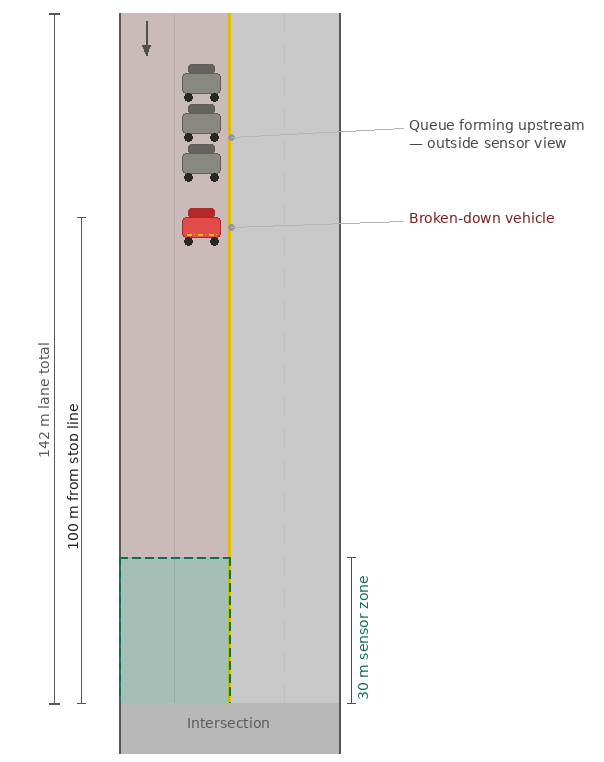}
\caption{Lane-level view of the disruption. The blockage is located upstream of the sensor zone, causing a queue that remains outside the observed region.}
\label{fig:lane_disruption}
\end{figure}

\paragraph{Scenario design.}
At $t=600$ (step~120), we introduce a disruption by placing a stationary vehicle on the northbound lane upstream of the sensor coverage region (Figure~\ref{fig:lane_disruption}). This causes a queue to form outside the observed region, making the resulting congestion unobservable in the agent’s input. The disruption lasts 600 seconds (steps 120--240), allowing congestion to accumulate and its long-term impact to become apparent.

In SUMO-RL, traffic is measured only near the stop line. Since the disabled vehicle is positioned upstream, the resulting queue lies outside the sensor’s field of view. Consequently, the corresponding queue feature remains \emph{light} throughout the accident window, even as congestion builds upstream.

This limitation also applies to the SPOT tree: each node is evaluated using the same sensor-based observation, so the blocked lane appears \emph{light} at all depths. SPOT does not detect the disruption directly, but instead captures its downstream effects.

\begin{figure}[h]
\centering
\includegraphics[width=0.9\linewidth]{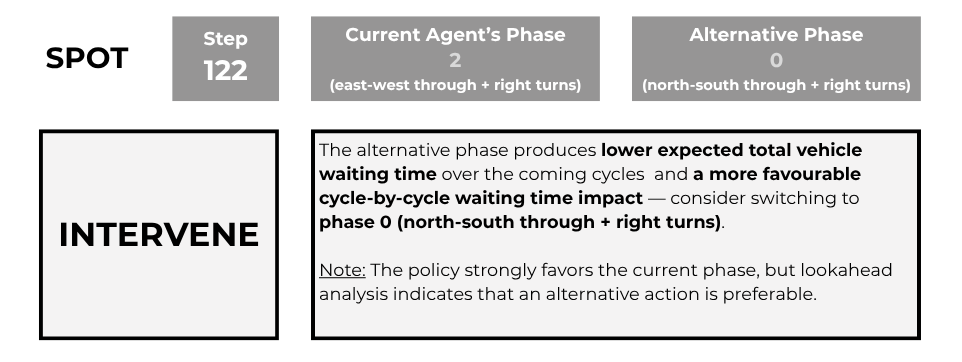}
\caption{
SPOT control panel for step~122. The system recommends \textsc{intervene}, suggesting a switch to an alternative phase that improves expected future traffic conditions.
}
\label{fig:control_panel}
\end{figure}

\paragraph{Example SPOT Explanation.}
To illustrate SPOT explanations, Figure~\ref{fig:control_panel} shows the control panel output for a representative decision during the accident window (step~122). Despite the agent strongly favoring the current phase, SPOT recommends \textsc{Intervene}, identifying an alternative phase that reduces expected future waiting time. Notably, this recommendation arises even though the observed state appears normal, highlighting SPOT’s ability to reason beyond immediate observations.

\begin{figure}[h]
\centering
\includegraphics[width=0.9\linewidth]{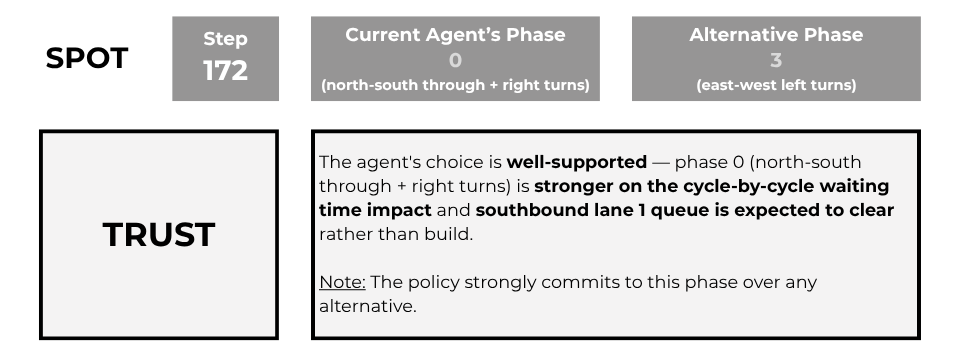}
\caption{
SPOT control panel for step~172. The system recommends \textsc{Trust}, indicating that the agent’s selected phase is supported by favorable expected future outcomes.
}
\label{fig:control_panel_172}
\end{figure}

In contrast, Figure~\ref{fig:control_panel_172} shows a later decision (step~172), where SPOT recommends \textsc{Trust}, indicating that the agent’s selected phase is well-supported. Here, the lookahead analysis confirms that the current action leads to favorable downstream outcomes.

\begin{figure}[h]
\centering
\includegraphics[width=0.8\linewidth]{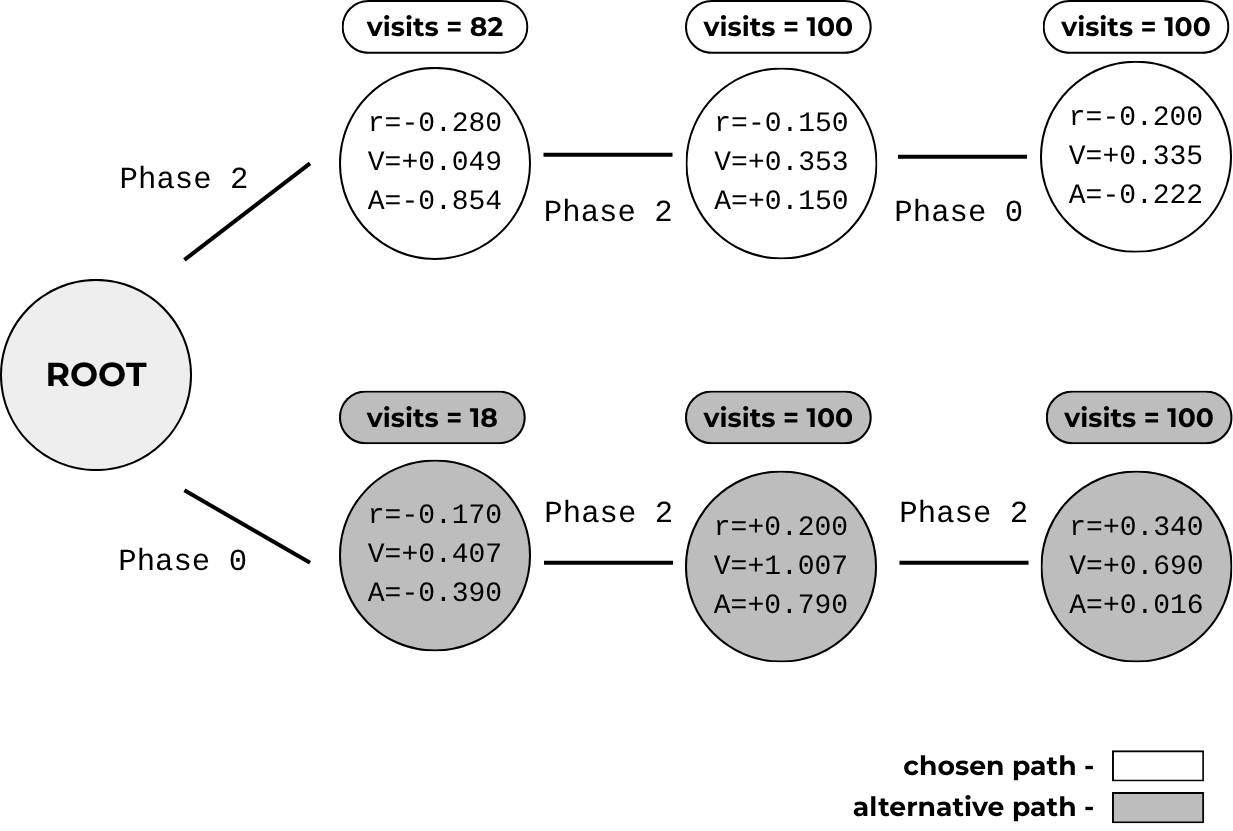}
\caption{
SPOT for step~122.
}
\label{fig:spot_tree}
\end{figure}

Figure~\ref{fig:spot_tree} shows the SPOT tree corresponding to step~122, where each node encodes visit counts, rewards, critic values, and advantages. The tree reveals that the alternative action leads to more favorable downstream outcomes, even when immediate signals are inconclusive.

\begin{figure}[h]
\centering
\includegraphics[width=0.8\linewidth]{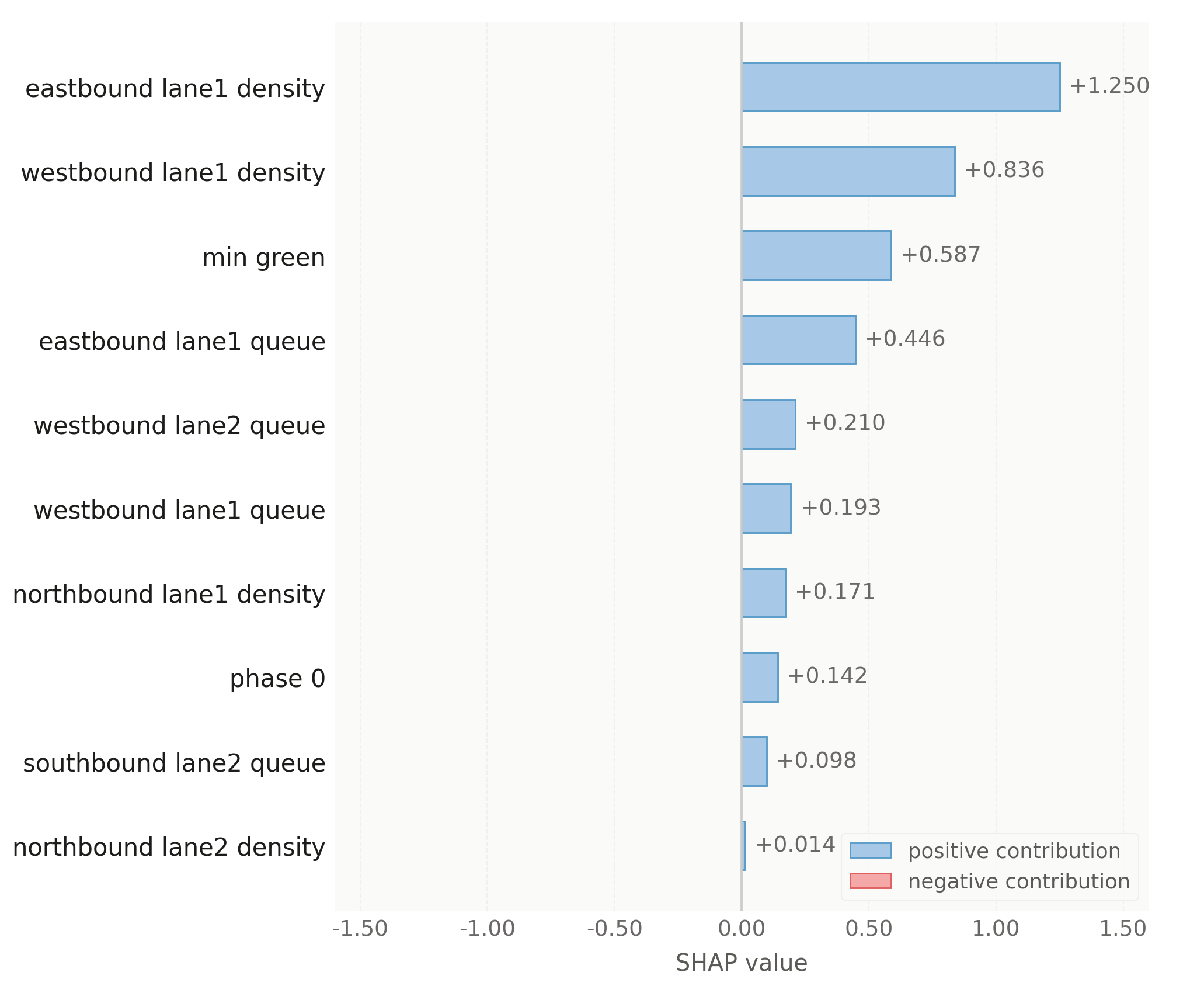}
\caption{
SHAP feature attribution for step~122.
}
\label{fig:shap}
\end{figure}

For comparison, Figure~\ref{fig:shap} shows feature-level explanations obtained using SHAP. 
Although SHAP identifies which features drive the agent’s immediate decision, it provides no information about the consequences of that decision. 
Because it operates on a single observation, SHAP cannot account for how traffic conditions evolve over time or how alternative actions would affect future states. 
In the presence of the upstream disruption—unobservable in the current feature vector—SHAP fails to detect the emerging congestion and instead explains a decision that appears locally reasonable.

\section{Conclusion}

We introduced \textbf{SPOT}, an explainable AI framework for deep reinforcement learning that constructs a sampling-based tree to capture the distribution of future trajectories induced by a policy. The approach is model-agnostic and applies across value-based, policy-based, and actor-critic methods.

We provided theoretical guarantees showing that SPOT faithfully reflects the underlying policy. Empirically, we demonstrated SPOT in a SUMO-RL traffic control setting, where it reveals behaviors not captured by single-timestep explanations. In particular, SPOT uncovers robustness issues arising from partially observable disruptions by reasoning over future trajectories, enabling more informed intervention in a human-in-the-loop setting.

As future work, we plan to conduct user studies to evaluate the actionability of SPOT explanations and their impact on human decision-making.

\clearpage

\bibliographystyle{named}
\bibliography{sample}

\end{document}